\documentclass[sigconf]{acmart}

\usepackage[english]{babel}
\usepackage{blindtext}
\usepackage{booktabs} 
\usepackage{amsmath}
\usepackage{epstopdf}
\usepackage{comment}
\usepackage{url}
\usepackage{color}

\renewcommand\footnotetextcopyrightpermission[1]{} 
\setcopyright{none}

\acmDOI{}

\acmISBN{}

\acmArticle{4}
\acmPrice{15.00}

\begin{document}
\title[Time Attention based Fraud Detection]{A Time Attention based Fraud Transaction Detection Framework}

%
%
%
%
%
%

\author[Longfei Li and others]{ Longfei Li$^\dag$, Ziqi Liu$^\dag$,  Chaochao Chen$^\dag$,Jun Zhou$^\dag$, Xiaolong Li$^\dag$}
\affiliation{%
	\institution{$^\dag$Ant Financial Services Group, Hangzhou, China}
}
\email{{longyao.llf,kaiyuan.lzq,chaochao.ccc,jun.zhoujun,xl.li,}@antfin.com}

%

\begin{abstract}
With online payment platforms being ubiquitous and important, fraud transaction detection has become the key for such platforms, to ensure user account safety and platform security. 
In this work, we present a novel method for detecting fraud transactions by leveraging patterns from both users' static profiles and users' dynamic behaviors in a unified framework. 
To address and explore the information of users' behaviors in continuous time spaces, we propose to use \emph{time attention based recurrent layers} to embed the detailed information of the time interval, such as the durations of specific actions, time differences between different actions and sequential behavior patterns,etc., in the same latent space. 
We further combine the learned embeddings and users' static profiles altogether in a unified framework. 
Extensive experiments validate the effectiveness of our proposed methods over state-of-the-art methods on various evaluation metrics, especially on \emph{recall at top percent} which is an important metric for measuring the balance between service experiences and risk of potential losses.

\end{abstract}

\keywords{Time Attention, Fraud transaction, Sequence, RNN}

\maketitle
	
\section{Introduction}

Online payment platforms have been playing an increasingly important role in our daily life, as we are heading towards a cashless society\footnote{\url{https://en.wikipedia.org/wiki/Cashless_society}}. The major online payment platforms, such as Alipay\footnote{\url{https://intl.alipay.com/}}, PayPal\footnote{\url{https://www.paypal.com}}  and Paytm\footnote{\url{https://paytm.com/}}, are currently serving hundreds of millions of users around the world and processing millions of cashless transactions each day. 
To provide a credible service,  a crucial and challenging issue is to ensure the safety of all the transactions, among which the detection and prevention for the fraud transactions is a critical task.

To handle this task, a key issue is how to construct the detection system. In recent years, machine learning based methods have been applied, in which the detection of fraud transaction is formulated as a classification problem and a model is trained with the collected labeled data ~\cite{panigrahi2009credit, mahmoudi2015detecting,srivastava2008credit}. 
When deployed, a score can be obtained for each transaction to measure the fraud risk with the trained model. 
Then a threshold is set so that those transactions whose scores are higher than the threshold will be suspended for further verifications, which include different authentication methods, such as face recognition, Short Messaging Services (SMS) and verification emails.
However, these are some awkward problems. 


When building a model, another important issue  is that the features in this task are much complicated, and specific consideration and a more effective model is needed. 
Roughly speaking, there are two different kinds of features in this task.
On the one hand, the users' static profiles, such as users' demographics and average spendings, are basic features to describe one user, and to indicate the risk of the account.
Thus, we claim that the model should pay enough attention to the dynamic features, and effective method should be explored to reduce the expense of computing and storage.

To handle sequence data, deep learning based methods, such as long short-term memory (LSTM)~\cite{DBLP:journals/neco/HochreiterS97}, convolutional neural network(CNN)~\cite{lecun1998gradient} , and their variant algorithms~\cite{DBLP:journals/corr/ChungGCB14}  have been developed in recent years, and significant improvement has been achieved in various applications, such as speech recognition~\cite{DBLP:conf/icassp/GravesMH13}, natural language processing~\cite{DBLP:conf/interspeech/MikolovKBCK10}, video processing~\cite{DBLP:conf/cvpr/DonahueHGRVDS15,DBLP:conf/icml/SrivastavaMS15}, etc. 
However, most of these methods only address the sequence information of the data, while the detailed information of the time intervals are not considered.
In fact, the time regularity is informative and the time interval information is meaningful in real-world applications. One example is that a person who trades in the same frequency is quite different from the one who trades irregularly. Another example is that the person who takes a short time between two operations is quite different from the one that takes a long time. 
Thus, the detailed information of the time interval is a key point which should be valued.

To address the problem above, we proposed to introduce the attention mechanism to handle it. 
Attention mechanisms have been proven to be a very powerful mechanism~\cite{DBLP:conf/nips/VaswaniSPUJGKP17}, and have brought improvement in many areas, such as natural language translation~\cite{DBLP:journals/corr/BahdanauCB14}, speech recognition~\cite{DBLP:conf/nips/ChorowskiBSCB15}.  The attention mechanisms are also applied upon CNN~\cite{DBLP:journals/tacl/YinSXZ16} or LSTM to integrate the extra sources of information, and guide the extraction of embeddings which are highly correlated to the specific tasks. 
As we discussed, detailed time information (not only the sequence information) is of great value, which will play crucial role in our task.
In this paper, inspired by~\cite{DBLP:journals/corr/LinFSYXZB17,DBLP:conf/nips/NeilPL16}, we propose to use \emph{time attention based recurrent layers} to embed the detailed information of the time interval, such as the durations of specific actions, time differences between different actions and sequential behavior patterns in the same latent space, etc., and we further combine the learned embeddings and users' static profiles altogether in a unified framework for the final training of the fraud transactions detection model.
The main contributions of this paper are summarized as follows.

\begin{itemize}
	\item We propose a novel time attention based recurrent layer which can operate sequential data in continuous time spaces with the detailed information of the time interval addressed.
	\item We perform experiments on several datasets, from which we conclude that our method is competitive and alternative to existing time-LSTM works.
	\item We further deploy the proposed framework  as a real system at Alipay, and the results on real-world tasks also validate the effectiveness of the proposed method in terms of various metrics.
\end{itemize}

The rest of this paper is organized as follows. We summarize the related literature in Section~\ref{related}, and
describe the detailed architecture of the proposed approach in Section~\ref{theory}. We report and discuss
experimental results in Section~\ref{exp}, and conclude in Section~\ref{conlusion}. 



\section{Background}
\label{related}
In this section, we will introduce the related literature that formed the basis of our work.

\subsection{Phased LSTM}

Phased LSTM ~\cite{DBLP:conf/nips/NeilPL16} is a RNN architecture for modeling event-based sequential data. It extends LSTM by adding the time gate $k_t \in \mathbb{R}$. $k_t$ is controlled by three parameters: $\tau$, $r_{on}$ and $s$, where $\tau$ is the total period of the model, $s$ is the phase shift and $r_{on}$ is the ratio which controls the ratio of the duration of the open phase to the full period. $\tau$, $r_{on}$ and $s$ are learned during the training process. $k_t$ is formally defined as:
\begin{eqnarray}
\phi_m &=& \frac{(t -s) \mod \tau }{\tau} \\
k_t &=& \left\{
\begin{aligned}
\frac{2\phi_t}{r_{on}} & \quad \text{if} \,\,\, \phi_t < \frac{1}{2}r_{on}, \\
2 - \frac{2\phi_t}{r_{on}} & \quad \text{if} \,\,\, \frac{1}{2}r_{on} < \phi_t < r_{ot}, \\
\alpha\ \phi_m, & \quad \text{otherwise},
\end{aligned}
\right.
\end{eqnarray}
where $t$ is the time stamp and $\phi_t$ is an auxiliary variable. And the modified model can be described as follows:
\begin{equation}
\begin{aligned}
\hat{c_t}  = & f_t \odot c_{t-1} + i_t \odot \delta_c(W_{xc}\ x_t\\
&+W_{hc}\ h_{t-1} + b+{c})  
\end{aligned} 
\end{equation}

\begin{eqnarray}
c_t &=& k_t \odot \hat{c_t} +(1 - k_t) \odot  c_{t-1} \\
\hat{h_t} &=& o_t \odot \delta_h( \hat c_t) \\
h_t &=& k_t \odot \hat h_t +(1 - k_t)\odot h_{t - 1}
\end{eqnarray}
where $x_t \in \mathbb{R}^d$ denotes input features at timestamps $t$, $h_t \in \mathbb{R}^k$ denotes
the $k$-dimensional hidden units, and $c_t \in \mathbb{R}^{k}$ denotes the cell memory.
However, this method is designed for high-frequency sampling scenes, which is quite different from our task.  

\subsection{Time LSTM}
Time LSTM ~\cite{DBLP:conf/ijcai/ZhuLLWGLC17} adds specific inner gated units in LSTM to maintain
the long term and short term effects on current actions in the sequence, such gates are controlled by the time interval between two actions. The model can be described as follows:

\begin{equation}
\begin{aligned}
T_m = &\sigma(x_m W_{xt} + \sigma_{\Delta t}(\Delta t_m W_{tt})+b_{t}),  \\
c_m  = & f_m \odot c_{m-1} \\
& + i_m \odot T_m \odot \sigma_{c}(x_mW_{xc} +h_{m-1} W_{hc} +b_c), \\
o_m = & \sigma_o(x_mW_{xo} \\ 
& + \Delta t_m W_{to} +h_{m-1}W_{ho} +w_{co} \odot c_m +b_o),
\end{aligned}
\end{equation}
where $\Delta t_m$ is the time interval between two states. Such a method has been successfully applied in
predicting users' next actions in recommendation systems (RS), which is quite similar to our task.
As mentioned in ~\cite{DBLP:journals/corr/BahdanauCB14},
by using the last hidden state of such models as the representation of the sequence, it's difficult to
use a fixed length vector to represent a long sequence.

\section{Time attention based Heterogeneous Network}
\label{theory}
The proposed heterogeneous network's architecture is shown in  Fig~\ref{fig:arch}. In this section, we will introduce the components of our architecture.

\subsection{Representation of Static Features}
\label{theory:raw}
 According to one's transaction and shopping records in the platform, we can collect one's profiles, such as working places, living places, credit scores (similar with FICO score\footnote{http://www.fico.com/}), trading amounts, etc., which demonstrate a person's consuming ability and habits. The rationale for using such features is that an unusual transaction amount or location may be suspicious.
 
As many continuous features  are static ones, before feeding such features into a neural network, data preprocessing, such as normalization and discretization, are needed. For example, different normalization, discretization methods are needed. But for tree-based models, the raw features can be directly used, as the model is able to split the numerical values accordingly. This property and the strong representation power of tree-based
 models make them widely adopted in the industries.
 Despite this, RandomForest(RF) or Gradient Boosting Decision Tree(GBDT) is a linear combination of separate trees, which can be observed from Eq.~\eqref{eq:boost},
 \begin{equation}
 F(x) = \sum_{i=1}^{n}\gamma_i h_i(x) + \text{const},
 \label{eq:boost}
 \end{equation}
 where $\gamma_{i}$ is the weight of the $i$-th tree, and $h_i(x)$ is the output of $i$-th tree.
 
 The boosted decision trees have shown to be a powerful model to transform the original features of an instance~\cite{DBLP:conf/kdd/HePJXLXSAHBC14}, which can then be utilized by other models to further get even higher accuracy. Specifically, we use each learned individual tree as a categorical feature, where the value is set as the index of the leaf node the instance falls in.
 As a result, if there are $n$ trees in the GBDT model, the transformed feature of an instance is given in terms of a structured vector $ x = (e_{i_1}, . . . , e_{i_n}) $,
 where $e_{i_k}$ is the $i_k$-th unit vector with the dimension of $d_k$, where $d_k$ is the number of leaf nodes at $k$-th tree, and $i_k$ is the index of the leaf node where the current instance falls into at $k$-th tree. 
 
\subsection{Dynamic Behaviors}
\label{theory:dynamic}
\subsubsection{Click Behavior}
\label{theory:click}
When users use services provided by Alipay, there will be a record describing the service the user had used, which is quite similar to the click history used in the recommendation system. We can formulate the user behavior sequence as a tuple $(u_j, a_i, t_i)$, where the $U=\{u_1,\dots,u_j,\dots, u_{|U|}\}$ is the user set, $A=\{a_1, \dots, a_i, \dots, a_{|A|}\}$ is the action set, $T=\{t_1, \dots t_i, \dots, t_{|T|}\}$ is the time stamp of user $u_j$ have done the action $a_i$. For a user $u$, his/her click behaviors can be represent as $B = \{(a_i, t_i) | i = 1, \dots m\}$. In order to involve the time effect, we separate the click behaviors into two parts, the first part is the click history $B_h = \{a_i | i = 1, \dots, m \}$, and the second part is the time behavior $B_t = \{t_i| i=1, \dots, m\}$. For the time behavior, we pay more attention to the interval between two actions, so we transform the time behavior as $B_t = \{\Delta t_1, \dots, \Delta t_{i})\}$, where $\Delta t_i= t_i - t_{i-1}$. However, since the $B_t$'s values fall into a large range, some values appear rarely, which makes the network hard to convergence, so a discretization process is needed.

\subsubsection{Transaction Behavior}
\label{theory:event}
When users make a transaction, a lot of information will be saved, which contain abundant aspects of this transaction, for example, an event will contain the scene, the location, and the time user does such transaction, at the same time the formal transaction place and the registered place are included in the event, which can demonstrate if the user is trading in an abnormal place. For time data, we use the same notation as mentioned in Section ~\ref{theory:click}.


\subsection{Time Attention based Recurrent Layers}
Since our attention mechanism is added upon RNN layers, so we will introduce the basic LSTM and GRU first, and followed by our proposed time attention mechanism.
\subsubsection{LSTM}
\label{lstm}
Using LSTM to model sequential data has many successful applications. Compared with Recurrent Neural Network(RNN), LSTM is comprised of forget gate, input gate, output gate, and a memory cell. Standard LSTM equations can be described as follow:
\begin{eqnarray}
i_t &=& \delta(W_i*x_t + U_i*h_{t-1} + b_i), \\
f_t &=& \delta(W_f*x_t + U_f*h_{t-1} + b_f)), \\
o_t &=& \delta(W_c*x_t + U_o*h_{t-1} + b_c), \\
g_t &=& \phi(W_g*x_t + U_g*h_{t-1} + b_g), \\
m_t &=& f_t \odot m_{t-1} + i_t \odot g_t, \\
h_t &=& o_t \odot \phi(m_t),
\end{eqnarray}
where the $W$, $U$ and $b$ are parameters of the LSTM. $x_t$ represents the input vector of the LSTM at  timestamps $t$, $\delta$ is the sigmoid function, $\phi$ is the hyperbolic tangent function.

%
%
%

\subsubsection{Time Attention}

Assuming we have a sequence consists of $n$ actions, represented in a sequence of embeddings:
\begin{equation}
    S = (\mathbf{x_1}, \mathbf{x_2}, \dots, \mathbf{x_n}),
\end{equation}
where $\mathbf{x_i}$ is a vector in dimension $d$, $S$ is a 2-D matrix, whose is $n$-by-$d$. The hidden states of RNN at time $i$ can be given by:

\begin{equation}
    h_i = \mathrm{RNN}(x_i, h_{i-1}),
\end{equation} 
where $\mathbf{h_i}$ is a $k$ dimension vector, $k$ is the hidden unit number of the RNN. All the n $h_i$s are denoted as $H$, whose shape is $n$-by-$k$ when RNN is the single direction architecture, or $n$-by-$2k$ if of a bi-direction architecture.
 
For time data, there are multiple meanings, for example, how long a user stays in a session, which means the degree of interest or familiar of this user, or how long after the user uses another service, which can denote a user's behavior. Here we use $\tau$ to denote the time data. Since $\tau \in \mathbb{R} $, where $\mathbb{R}$ is one dimension real value space, we first discrete the time data by $\theta = \biggl\lfloor\frac{\tau}{60}\biggr\rfloor$, and just use it as a category feature, and then we can encode the time data as:
\begin{equation}
    \rho_{i} = \mathrm{lookup}(\theta_i),
\end{equation}
 where $\rho_{i}$ is the embedding representation of the discrete time data, whose dimension is $g$. Then we can stack all time embeddings $\{\rho_1, \dots \rho_n \}$ together, and denote such matrix as $P$, which means the embedding representation of time data. Its dimension is $n$-by-$g$. Following the self-attention mechanism, we use the following equations to calculate the weight of $H$ we get from RNN:
\begin{eqnarray}
\epsilon &=& \mathbf{w_s} \tanh(W_e*H^{\top} + W_t*P^{\top}), \\
\mathbf{\alpha} &=& \mathrm{Softmax}(\epsilon),
\end{eqnarray}
where $W_e$ and $W_t$ are the matrices to be learned,  $W_e \in \mathbb{R}^{m \times k}$,  $W_t \in \mathbb{R}^ {m \times g}$, $\mathbf{w_s}$ is a vector, $\mathbf{w_s} \in \mathbb{R}^{m}$. $\alpha$ is the attention weight which quantifies the relevance of features in $H$. The $\mathrm{Softmax}$ ensures all the computed weights sum up to 1. After getting the $\mathbf{\alpha}$, we can use the standard attention mechanism to gather the embeddings extracted from different time states together by the following equation:

\begin{equation}
\hat{h} = \sum_{i=1}^{n} \alpha_i * h_i ,
\end{equation}
where the $\hat{h}$ demonstrates the new representation of the sequence.

\begin{figure*}
    
    \centering
    \includegraphics[height=3.5in, width=6.5in]{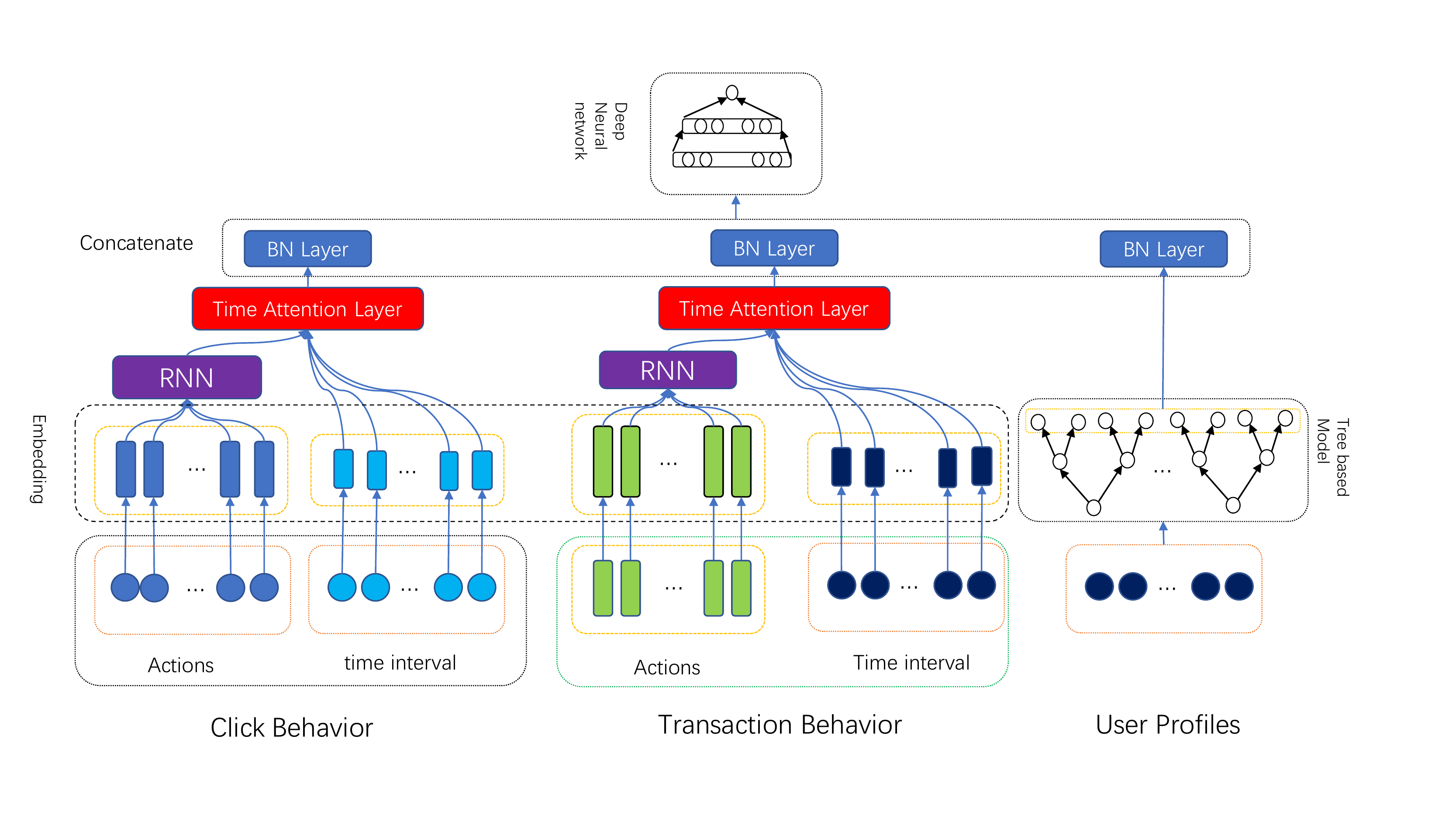}
    \caption{Architecture of our framework.}
    \label{fig:arch}
\end{figure*}

\subsection{Heterogeneous network}
Since we have two different kinds of behavior data, and static features in our system, we want to blend the heterogeneous data into a unified architecture, which will make the whole system more compact, at the mean while reduce the work of feature engineering. According to the method we mentioned in Section ~\ref{theory:raw}, we extract tree embeddings based on the user profile. At the same time, we extract two kinds of behavior embeddings from click and transaction behavior by our time attention RNN architecture, respectively. Since the values from different parts are in different scales, directly concatenating them together will make the whole network hard to converge. So we add a batch normalization layer~\cite{DBLP:conf/icml/IoffeS15} at the top of time attention layer and tree embedding layer, then we concatenate the output of each BN layer and feed them to a multi-layer neural network. The whole architecture is shown in the Figure~\ref{fig:arch}.
\section{experiments}
\label{exp}
In this section, we will describe the comprehensive experiments that we conducted to show the effectiveness of our proposed model. 
We first describe the dataset we use, the comparison methods, hyper-parameter settings, and evaluation metrics. We then report the comparison result and finally study model parameter effects. 

\subsection{Dataset}
We use the real transaction data from Alipay as our experimental dataset, where both real and fraud transactions are available. 
Fraud detection task is quite different from the traditional classification tasks, because the execution methods of fraud transaction vary in different time periods. 
Thus, in order to test our model's performance as practical as possible, we separate the original five-month transaction data into three parts according to the transaction occur time: 
the transaction data of the first three months are used as training set, the data of the fourth month is used as the validation set, and the data of the last month is used as test set. 
Meanwhile, since the whole transaction amount is extremely large and the fraud transactions are rare, we down-sample the non fraud samples of train set and the validation set to accelerate our experiments, at the same time, in order to simulate the real online situation, we sample from the original data set to build our test set which makes the fraud and non-fraud samples' number is quite different from train and validation set.
We report the details of the dataset after preprocessing in Table ~\ref{tab:fraud_dataset}. 

\begin{table*}[htpb]
    \caption{Fraud detection dataset description}
    \label{tab:fraud_dataset}
    \begin{tabular}{|c|c|c|c|c|}
        \hline
        Dataset & \#User  & \#Sequences & \#Non Fraud Transaction & \#Fraud Transaction  \\
        \hline
         train set & 1,221,706 & 3,837,624  & 3,832,560 &5,064\\
        \hline
         validation set& 656,521 & 1,248,912 & 1,247,315& 1,597  \\
        \hline
        test set& 674,057 & 1,302,226 &1,302,091 & 135 \\
        \hline
    \end{tabular}
    \vspace{-0.05in}
\end{table*}

As described in section~\ref{theory}, the features in fraud transaction scenario are divided into three groups, i.e., user static features, user click behavior features, and transaction behavior features. 
User static features demonstrate a person's consuming ability and habits. 
For the click behavior, we choose user's interactions with the Alipay APP during the recent two days as click behavior features. 
We set the max number of interaction to 200 based on experience. If the number of interaction is bigger than 200, we only keep the lasted 200 interactions. 
For transaction behavior, user's trading history in Alipay during the recent ten days are selected as features.
We also set the max number of trading history to 32 based on experience. If the number of history is bigger than 32, we will only keep the latest 32 transactions.
Moreover, for each transaction, we select the 28 most important attributes, e.g., the trading location, IP location, trading amount and so on. Finally, we summarize the dimensions of each feature in Table \ref{tab:feature dimension}.

\begin{table}[htpb]
    \caption{Feature dimension description}
    \label{tab:feature dimension}
    \begin{tabular}{|l|l|}
        \hline
        Static feature dimension  & 89 \\
        \hline
        User Behavior feature dimension  & 2,300\\
        \hline
        Transaction feature dimension & 28 \\
        \hline
    \end{tabular}
\end{table}


\subsection{Comparison Methods}
\label{compare}
In order to study whether our proposed time attention mechanism works, we compare our time attention mechanism with the following methods by varying the building block that generates the behavior embedding.

\begin{itemize}
    \item Bi-LSTM: We use bidirectional-LSTM ~\cite{graves2009offline} method to model the user's behavior. We extract the last state data as user embedding, and concatenate it with tree embedding extracted from trees~\cite{DBLP:conf/www/ZhouCLZQH17},~\cite{DBLP:journals/tist/ZhangZZFLLLZCLQ19}.
    \item Phased LSTM: This method is introduced in Section ~\ref{related}. We use the implementation which is provided by TensorFlow ~\cite{DBLP:journals/corr/AbadiABBCCCDDDG16}.
    \item Self-attention LSTM: We add a self attention layer on the top of Bi-LSTM which is introduced in ~\cite{DBLP:journals/corr/LinFSYXZB17}.
    \item CNN+Max pooling: We use traditional CNN with Max pooling to extract click and transaction behavior's embedding. The window size is set from 4 to 10. For click behavior, the kernel size is set to be 32. For transaction behavior, the kernel size is set to be 16, which equals to the embedding dimension of different kinds of behaviors.
    \item Time LSTM: This method is introduced in Section ~\ref{related}, and we use the implementation available at GitHub\footnote{https://github.com/DarryO/time lstm}.
    
\end{itemize}

\subsection{Hyperparameter Setting}
We fix the tree model's parameters, so that different models are using the same tree embeddings. 
For all the LSTM derived algorithm, we set the stack depth to 1, and use the same shape to make a fair comparison. 
The detailed settings are described as below. 
\begin{itemize}
    \item{Tree Embedding}: We choose the large-scale GBDT model implemented on KunPeng ~\cite{DBLP:conf/kdd/ZhouLZCLYCYCDQ17} as the tree model, and we set the tree number to 100 and the max deepth to 5.
    \item{Network shape}: For LSTM, GRU, and the derived algorithm, we set the hidden units to 256. For MLP, the hidden layer size is set to 1, and hidden unit number is 128.
    \item{Learning rate}: We use SGD as the optimizer, and select the best learning rate in \{0.1, 0.01, 0.001\} . 
    \item{Embedding Dimension}: For every time stamp, the transaction event contains 28 different features, and each feature contains a different number of components, each component uses a 16 dimension embedding matrix. For click behavior, the dimension of embedding matrix is set to  32. For the time dimension, we select the best value in \{8, 16, 32, 64\}.
    \item{Batch Size}: We set batch size to 512 for all the models.
    \item{Regularization}:     We use L2 as regularization, and its value is set to be 1e-5.
    
\end{itemize}

Note that for each model, we use the validation set to select the best model parameters, and evaluate them on the test set.

\subsection{Evaluation Metrics}

We use three different kinds of evaluation metrics to measure our proposed method's performance. We adopt two standard ranking metrics: Area Under ROC Curve (AUC) and F1-Score. At the same time, in the real fraud detection system, we can not disturb too many people to improve the recall rate, we use another more practical indicator to evaluate our method, i.e., \textbf{R}ecall \textbf{A}t \textbf{T}op \textbf{P}ercent (\textbf{RATP}). RATP@$r$ is the recall of the subset which consists of the instances of the top $r$ percentage of prediction scores, for example, RATP@0.05 means only 5 transactions will be disturbed in 10000 transactions.

\begin{table*}[htpb]
    \caption{Experiment Results}
    \label{tab:exp_result}
    \begin{tabular}{|c| c|c|c|c|}
        \hline
        Method & F1-score  & AUC & RAPT@0.05 & RATP@0.01 \\
        \hline
        GBDT & 0.701 & 0.981  &0.807&    0.637  \\
        \hline
        CNN+Max pooling &0.702&0.982&0.815&0.652 \\
        \hline
        GRU& 0.708 & 0.981 &0.822  &0.652 \\
        \hline
        Bi-LSTM& 0.712 & 0.983 &0.815& 0.659\\
        \hline
        Bi-LSTM+self attention& 0.714&0.984 &0.830 &0.674\\
        \hline
        PLSTM & 0.714& 0.986&0.835&0.689 \\
        \hline
        TLSTM&0.716& 0.986&0.844&0.692 \\
        \hline
        Bi-LSTM+time attention& \textbf{0.721}&\textbf{0.99} &\textbf{0.864}&\textbf{0.706} \\
        \hline
        GRU + time attention &0.718&0.988&0.859&0.703\\
        \hline
    \end{tabular}
\end{table*}


\begin{figure}
    \centering
    \includegraphics[height=140pt]{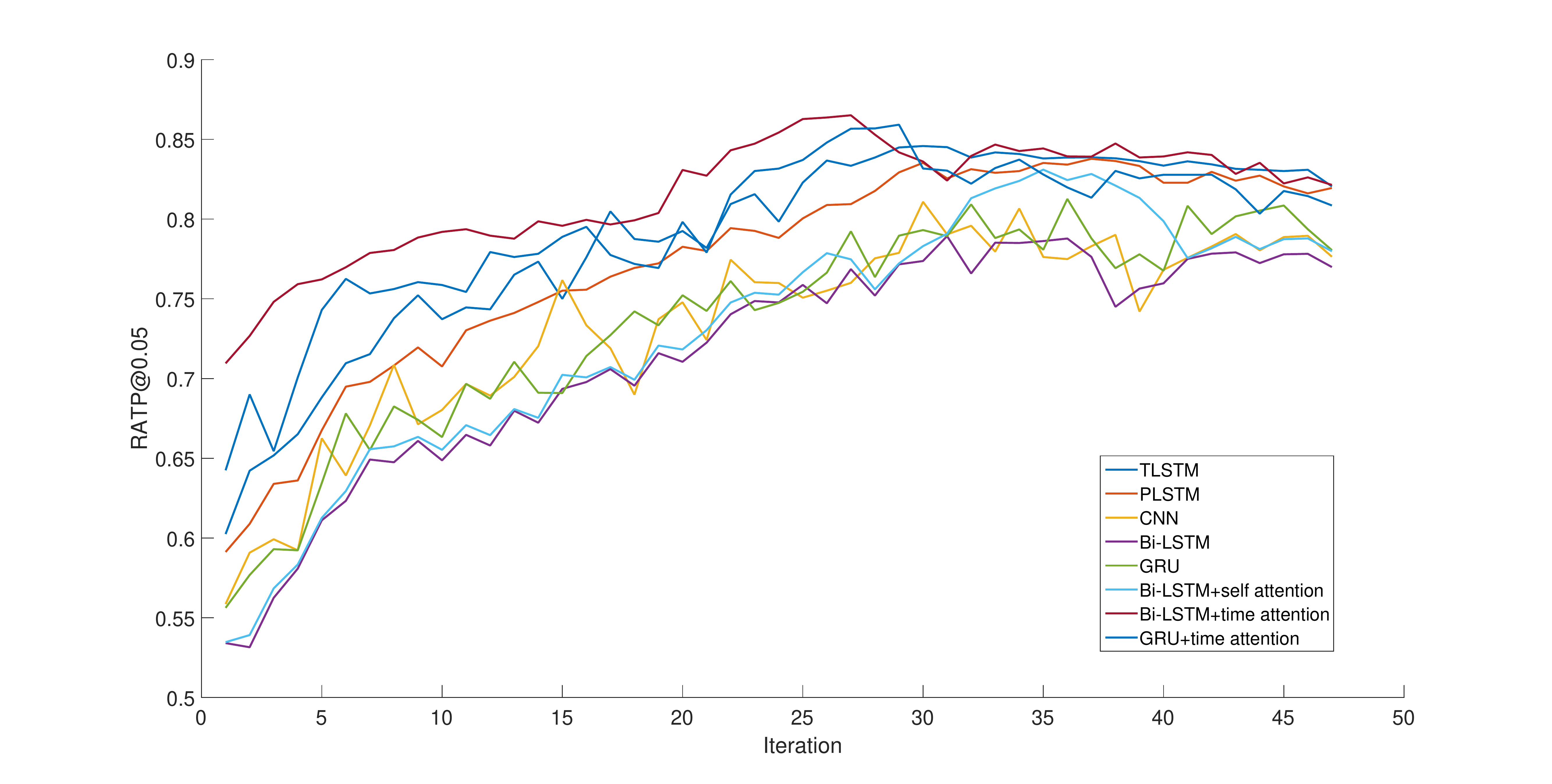}
    \caption{Recall@Top 0.05\% result at test set.The vertical axes denotes RATP@0.05 on the test set and the horizontal axes denotes training iterations. For each iteration, the model have processed 512*1000 sequence.}
    \label{fig:ratp5}
\end{figure}

\begin{figure}
    \centering
    \includegraphics[height=140pt]{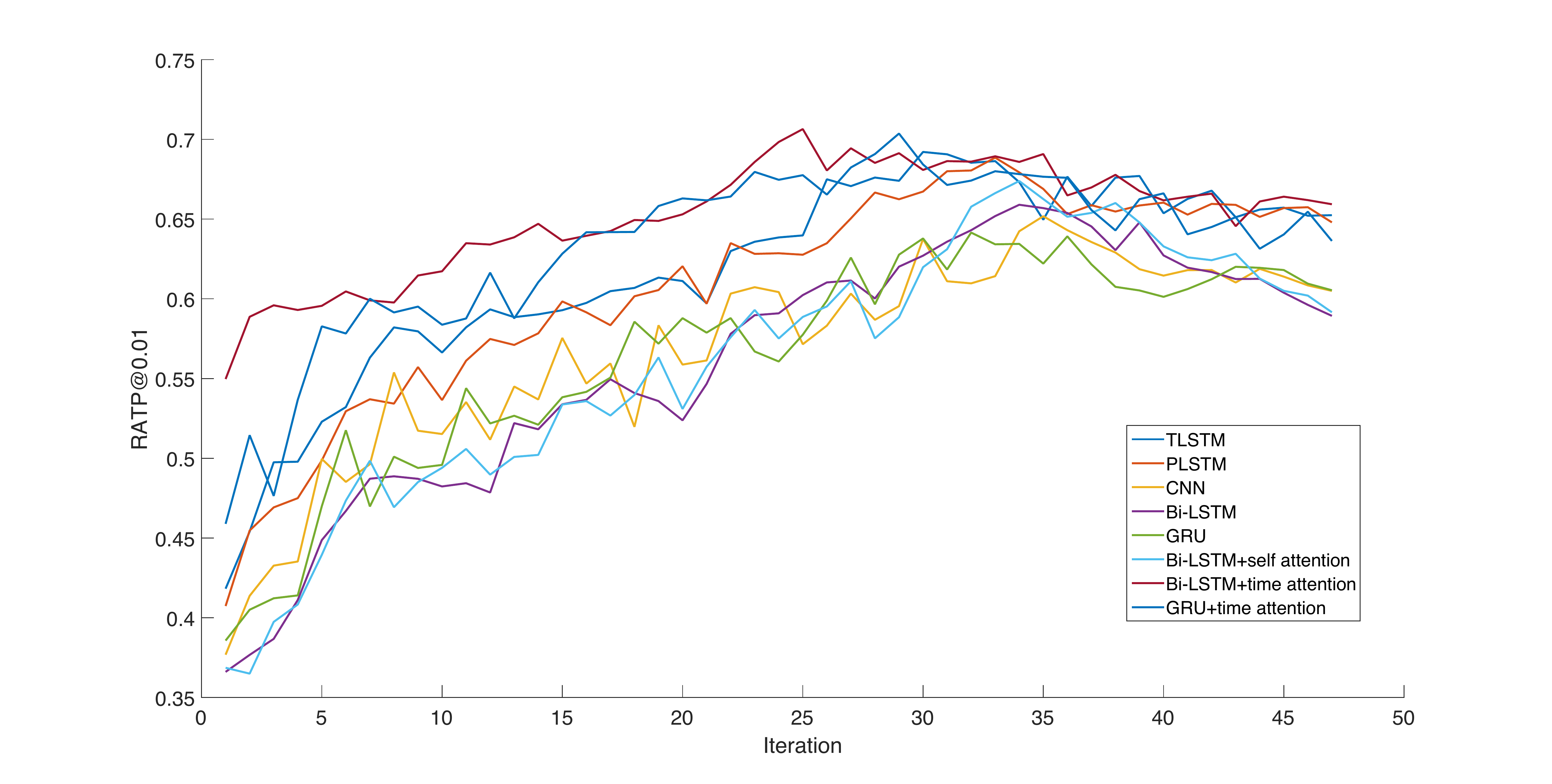}
    \caption{Recall@Top 0.01\% results at test set, whose meaning is similar as Fig ~\ref{fig:ratp5}}
    \label{fig:ratp1}
\end{figure}

\subsection{Comparison Results}
We report the comparison results in Table~\ref{tab:exp_result}.
From it, we can see that: (1) compared with the original GBDT which uses behavior features extracted by human, after using LSTM or GRU to modeling the user behavior, our proposed model has a significant improvement in terms of RATP@$r$.
Take RATP@0.05 for example, our proposed method has a 7\% improvement compared with the GBDT, which is because by using sequence modeling method, more complex patterns can be extracted. 
(2) The improvements of our proposed model against other models are not significant in terms of AUC, which is because the number of Non-Fraud transaction is too many, while the number of fraud transaction is too little. Thus, the improvement at the high score part will not improve the AUC too much. 
(3) All the methods that consider the time influence between different action outperform the Bi-LSTM and GRU and Bi-LSTM with self-attention, which means that time is an important information in fraud detection task. 
(4) At the same time, LSTM, GRU with our proposed time attention mechanism outperform PLSTM and TLSTM in our task. This is because, compared with adding inner gates in RNN, time attention mechanism that uses time information to guide the generation of sequence embedding may provide a better representation of the time sequence. 

We also show the convergence speed of different models in Fig ~\ref{fig:ratp5} and Fig ~\ref{fig:ratp1}. From them, we can find our proposed method's convergence speed is the fastest, and LSTM with time attention is slightly better than that of GRU with time attention. 

\subsection{Parameter Analysis}
We will study the effects of the hidden units and the time embedding dimension on our model performance. 

\subsubsection{Effect of the hidden units}
\begin{figure}
    \centering
    \includegraphics[height=150pt]{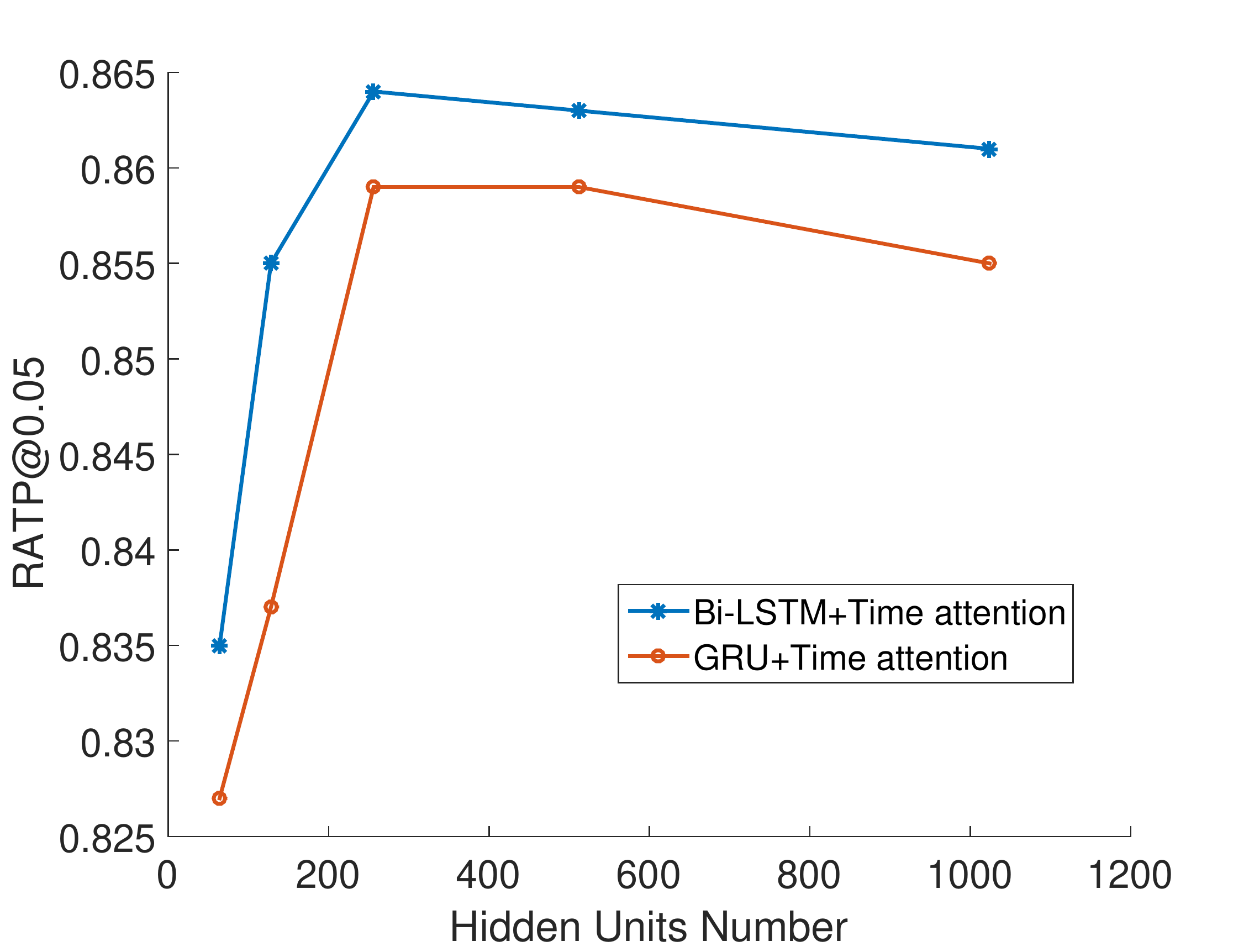}
    \caption{Effect of different hidden units. The vertical axes indicates test set RATP@0.05 and the horizontal axes indicates the number of RNN hidden units.}
    \label{fig:hidden}
\end{figure}

We first vary the LSTM/GRU hidden units number to study their effect on our model performance, while fixing other hyperparameters. 
The result is shown in Fig ~\ref{fig:hidden}. As we can see, with $n_h$ increases, the performance of RATP@0.01 becomes better. However, $n_h=\{1024, 512\}$ do not perform too much better than $n_h=256$. This is because as the number of hidden unit increasing, the parameter is also increasing, which makes more data is needed to fit the model.

\subsubsection{Effect of the time embedding dimension}

\begin{figure}
    \centering
    \includegraphics[height=150pt]{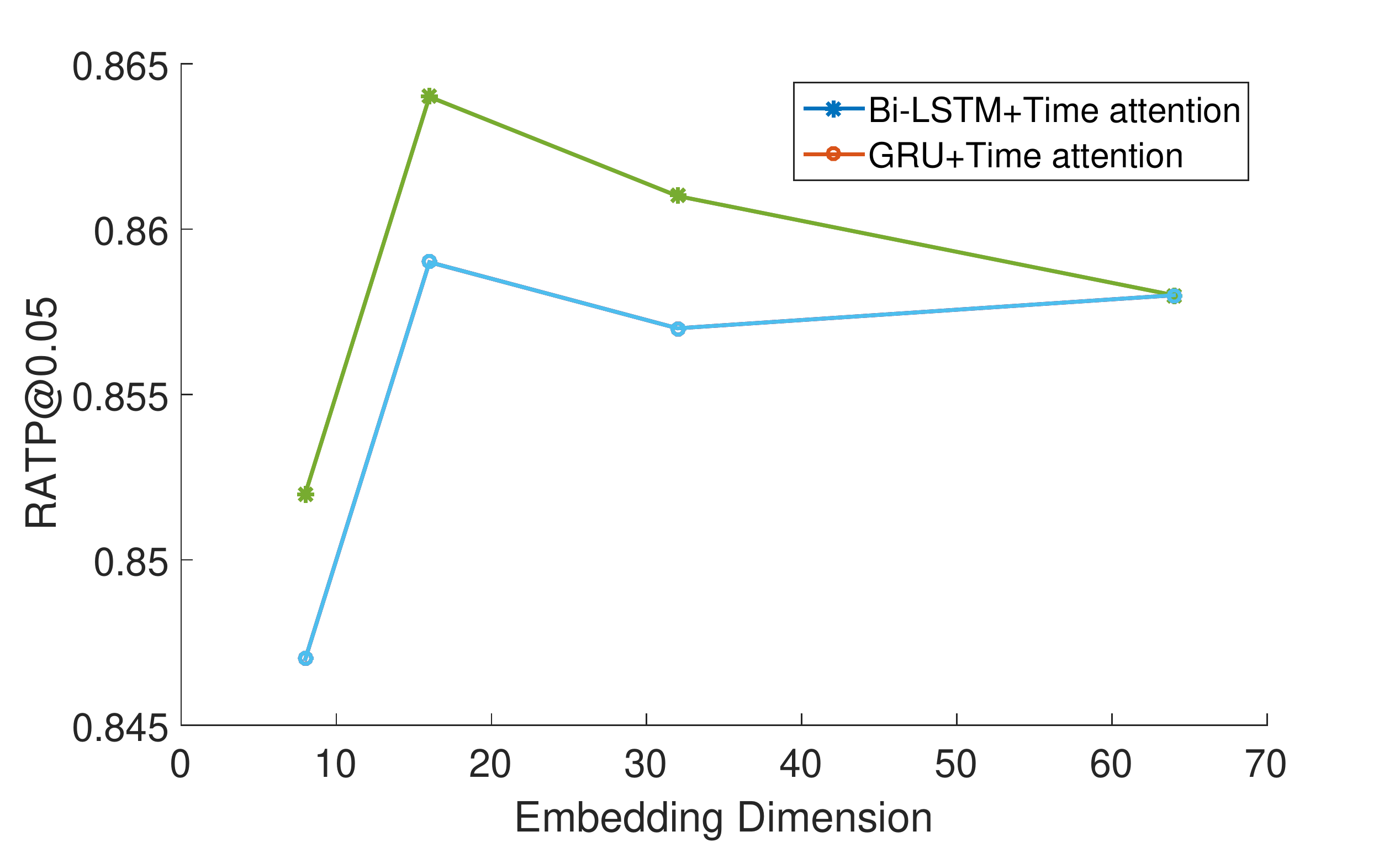}
    \caption{Effect of different time embedding dimension. The vertical axes shows RATP@0.05 on the test set and the horizontal axes is the dimension of time embedding. }
    \label{fig:time_embed}
\end{figure}

We then vary the time data's embedding dimension to study its effect on model performance, at the same time we fixing the other hyper-parameters. As shown in the Fig ~\ref{fig:time_embed}, with the time data embedding dimension increases, the performance does not always become better. When time embedding dimension is 32, we get the best result. That because as the time embedding dimension increasing, the feature space become sparse, which makes the model harder to converge.

%
%
%
%

\section{Conclusions and future work}
\label{conlusion}
In this paper, we proposed a framework which manipulates heterogeneous data, at the same time, we introduce a new attention mechanism which models the time aspect into the whole framework. We implemented and evaluated our proposed method against several baseline approaches, and showed that our method achieve the best results. 

In the future, we will try to evaluate our model in more datasets, and we will improve the computational efficiency of our model. Moreover, we will try to deal with the users who do not have too much history information in our platform.

\bibliographystyle{ACM-Reference-Format}

\bibliography{lstm} 

\end{document}